\documentclass[10pt,twocolumn,letterpaper]{article}

\usepackage[pagenumbers]{iccv} 

\usepackage{graphicx}
\usepackage{subcaption}
\usepackage[linesnumbered,ruled,vlined]{algorithm2e}
\usepackage{algorithmic}
\usepackage{amsmath}
\usepackage{amssymb}
\usepackage{multirow}
\usepackage{adjustbox}

\graphicspath{{./figure/}{../}}

\definecolor{iccvblue}{rgb}{0.21,0.49,0.74}
\usepackage[pagebackref,breaklinks,colorlinks,allcolors=iccvblue]{hyperref}

\usepackage{multirow}
\newcommand{\minisection}[1]{\vspace{.03in}\noindent{\textbf{#1}.}}
\usepackage{amsmath}

\newcommand*\wcircled[1]{{                  
    \large{\textcircled{\scriptsize{\textbf{#1}}}}
}}

\title{Does Your Vision-Language Model Get Lost in the Long Video Sampling Dilemma?}
\author{
Tianyuan Qu\textsuperscript{1}\footnotemark[1], Longxiang Tang\textsuperscript{2}$^*$, Bohao Peng\textsuperscript{1}, Senqiao Yang\textsuperscript{1}, Bei Yu\textsuperscript{1}, Jiaya Jia\textsuperscript{2} \vspace{.5em}  \\
\textsuperscript{1}CUHK \quad 
\textsuperscript{2}HKUST \\
}

\begin{document}
\twocolumn[{
    \vspace*{-0.3in}
    \renewcommand\twocolumn[1][]{#1}
    \maketitle
    \vspace*{-0.4in}
    \centering
    \captionsetup{type=figure}
    \hspace*{-0.35cm}
    \captionsetup{skip=-5pt}
    \includegraphics[width=1.05\textwidth]{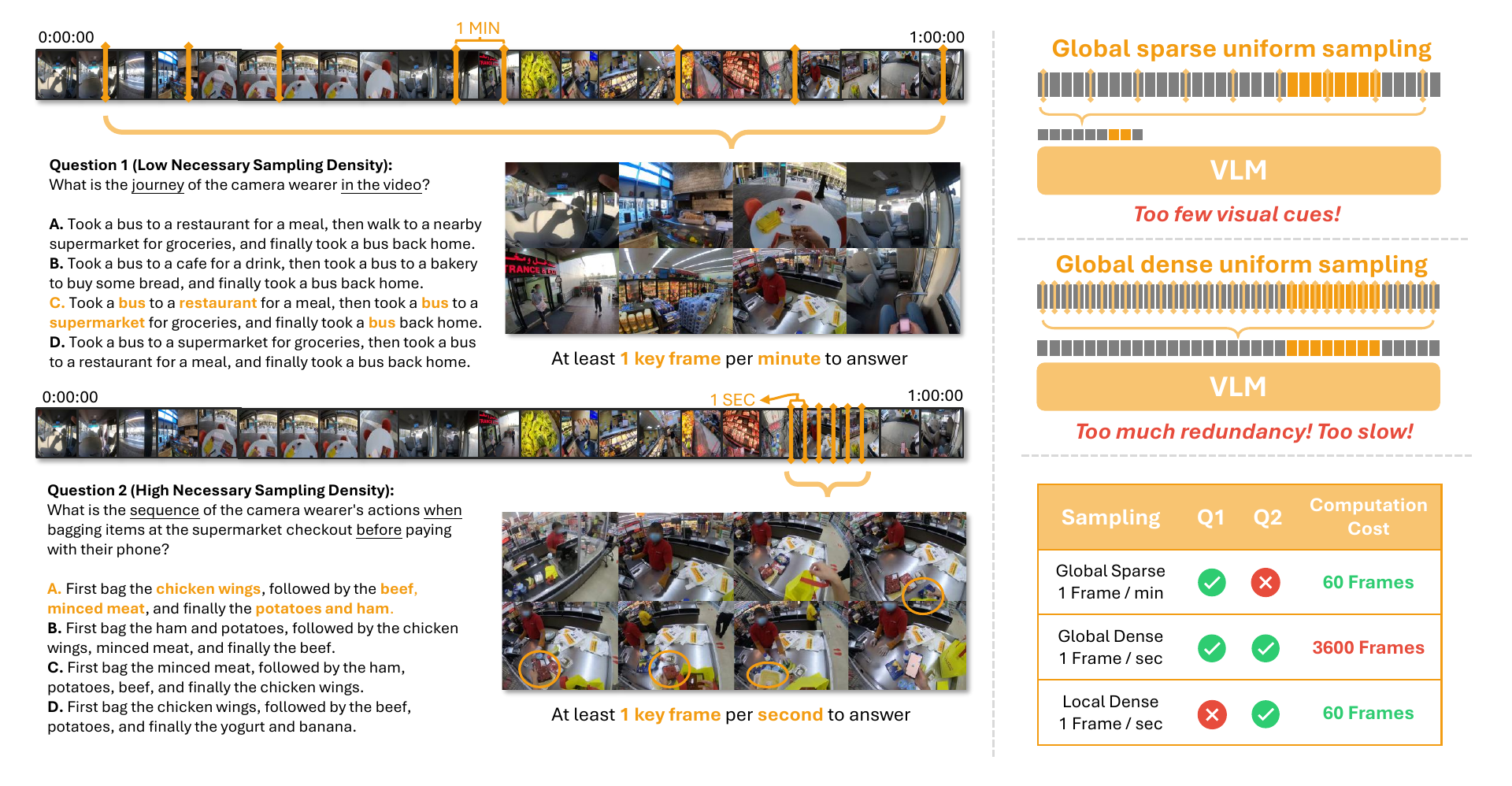}
    \caption{\textbf{Illustration of the Sampling Dilemma in Long Video Analysis:} (\textbf{Left}) In Q1, identifying a camera wearer's visited locations requires analyzing the entire video. However, key frames are sparse, so sampling one frame per minute often provides enough information. In contrast, Q2 examines the packing order during checkout, requiring high-resolution sampling to capture rapid actions. (\textbf{Right}) \textbf{Sampling Dilemma} emerges in tasks like Q2: a low sampling density fails to provide sufficient visual cues for accurate answers, while a high sampling density results in redundant frames, significantly slowing inference speed. This challenge underscores the need for adaptive sampling strategies, especially for tasks with high necessary sampling density.}
    \vspace{0.5cm}
    \label{fig:teaser}
}]

\renewcommand{\thefootnote}{\fnsymbol{footnote}}
\footnotetext[1]{Equal contribution.}
\renewcommand{\thefootnote}{\arabic{footnote}}

\begin{abstract}
The rise of Large Vision-Language Models (LVLMs) has significantly advanced video understanding. However, efficiently processing long videos remains a challenge due to the ``Sampling Dilemma'': low-density sampling risks missing critical information, while high-density sampling introduces redundancy. To address this issue, we introduce LSDBench, the first benchmark designed to evaluate LVLMs on long-video tasks by constructing high Necessary Sampling Density (NSD) questions—where NSD represents the minimum sampling density required to accurately answer a given question. LSDBench focuses on dense, short-duration actions to rigorously assess the sampling strategies employed by LVLMs. To tackle the challenges posed by high-NSD questions, we propose a novel Reasoning-Driven Hierarchical Sampling (RHS) framework, which combines global localization of question-relevant cues with local dense sampling for precise inference. Additionally, we develop a lightweight Semantic-Guided Frame Selector to prioritize informative frames, enabling RHS to achieve comparable or superior performance with significantly fewer sampled frames. Together, our LSDBench and RHS framework address the unique challenges of high-NSD long-video tasks, setting a new standard for evaluating and improving LVLMs in this domain. Our benchmark and evaluation codes has been released at: \href{https://github.com/dvlab-research/LSDBench}{https://github.com/dvlab-research/LSDBench}
\end{abstract} 
\section{Introduction}
\label{sec:intro}

With the emergence of large vision-language models (LVLMs)~\cite{bai2025qwen2,hurst2024gpt,team2024gemini,li2024llava,liu2023visual,chen2024internvl,alayrac2022flamingo,lin2023video,zhong2024lyra}, research on video understanding~\cite{li2024llava,bai2025qwen2,wang2025internvideo2,chen2024longvila,zhang2024long,zhang2023video,maaz2023video,song2024moviechat,li2024llama} has advanced significantly; however, tasks involving the understanding of long videos~\cite{fu2024video,wu2024longvideobench,mangalam2023egoschema,zhou2024mlvu} remain in an exploratory stage.
For LVLMs, learning from real-world recordings is vital to grasp the complexities of the physical world. This capability is particularly critical for downstream applications such as embodied intelligence~\cite{huang2020quality,huang2019comyco,wang2023mimicplay}, wearable devices, and security surveillance~\cite{niu2004human,zhou2019anomalynet}, where the ability to process long videos efficiently and accurately is a fundamental requirement.

To complete video tasks effectively, it is necessary to acquire a sufficient amount of visual cues relevant to the given question. Existing video LVLMs~\cite{bai2025qwen2,wang2025internvideo2,team2024gemini,lei2021less} typically utilize a global uniform sampling strategy to collect such visual cues. While this approach works well when videos are short or the required visual cues persist over a long segment, it encounters significant challenges when the critical visual cues are concentrated within a small portion of a long video. In such cases, increasing the sampling density may appear to be a solution. However, this introduces the ``\textbf{Sampling Dilemma}'', which arises from the following trade-offs:
\begin{itemize}
    \item With low sampling rates risk overlooking critical information in essential segments, leading to task failure.
    \item Conversely, with high sampling rates, numerous redundant frames from irrelevant segments are included, which wastes valuable context length, increases computational overhead, and potentially introduces noise, thereby reducing accuracy.
\end{itemize}

For example, consider Question 2 in \Cref{fig:teaser}, which asks about the details of packing items at a cashier counter.
The relevant actions occur within a brief segment of a one-hour video, with each step lasting about one second. Answering this question using global uniform sampling would require a high rate—around 1 frame per second—to capture sufficient cues. However, this approach leads to an overwhelming number of irrelevant and redundant frames, significantly increasing computational costs and often exceeding the input context length of VLMs. This trade-off between sampling density and efficiency is a core challenge for long-video tasks.

To emphasize the sampling dilemma in long video tasks, we first define the concept of \textbf{Necessary Sampling Density (NSD)}:
the minimum sampling density required, under uniform sampling, to extract sufficient information from a video to answer a question.
Building upon this, we introduce the \textbf{L}ong-video \textbf{S}ampling \textbf{D}ilemma \textbf{Bench}mark (\textbf{LSDBench}), which consists of a set of questions characterized by high NSD requirements and videos lasting for hour. In our benchmark, we construct questions about action sequences within short segment of long videos~\cite{grauman2022ego4d}, such as Question 2 in \Cref{fig:teaser}. 
These sequences involve densely distributed, short-duration actions, ensuring a high NSD for each question.
Meanwhile, we incorporate temporal referring in the question phrasing to explicitly direct attention to the target segment. This design—featuring local segments with high NSD and a high proportion of question-irrelevant content across the whole video—provides a more intuitive and rigorous way to evaluate the effectiveness of different sampling strategies.

To efficiently construct such a dataset, we developed a fully automated pipeline for question generation and filtering.
Central to this process is a temporal structuring strategy, designed to effectively organize video content for question generation.
To guarantee the quality of the generated questions, we introduced a two-round filtering mechanism, which successfully eliminated over 90\% of low-quality questions. 
Furthermore, we incorporated an adversarial iteration strategy to enhance the distractor options in multiple-choice questions.
This approach minimizes the possibility of models correctly answering questions based solely on the text of the question and options,
particularly in scenarios where the sampled visual cues are limited.

With LSDBench, we can investigate sampling dilemma quantitatively.
A straightforward method to tackle it is to allow the model to perform dense sampling only in important regions.
Based on this, we propose a two-stage \textbf{Reasoning-Driven Hierarchical Sampling (RHS)} framework: in the first stage, the video is sparsely sampled, and the VLM locate visual cues relevant to the question via reasoning. In the second stage, the segments corresponding to these key frames are densely sampled to provide sufficient information for the model to infer the final answer. Considering the potential missing of necessary information during sparse sampling, we further introduce a light-weighted \textbf{Semantic-guided Frame Selector (SGFS)}, extracting an optimal frame set with maximum visual information. Using this approach, our method enables existing models to achieve better performance than uniform global sampling, even with a significantly reduced number of sampled frames.

Our contributions can be summarized as following:
\begin{itemize}
    \item \textbf{LSDBench:} We propose LSDBench, a benchmark that focuses on the sampling dilemma in long-video tasks. Through well-designed tasks, it evaluates the sampling efficiency of long-video VLMs. We also provide a fully automated data generation and filtering pipeline, which can easily construct high-quality multiple-choice QA pairs for very long video.

    \item \textbf{Reasoning-Driven Hierarchical Sampling:} We propose a two-stage, training-free framework that improves long-video processing efficiency by focusing VLM on important segments. This provides a sampling-efficient solution for handling long video tasks.

    \item \textbf{Semantic-Guided Frame Selector:} We introduce a module that selects frames with higher visual information content without any question prior. Combined with RHS, this approach achieves comparable or even better accuracy on LSDBench with significantly fewer total sampled frames.
\end{itemize}

\section{Related Work}
\label{sec:relatedwork}

\minisection{Long Video Benchmarks}
In recent years, benchmarks for evaluating long-video processing capabilities~~\cite{fu2024video,zhou2024mlvu,mangalam2023egoschema,chen2024longvila,chandrasegaran2024hourvideo} have gained attention. EgoSchema~\cite{mangalam2023egoschema} and ActivityNet-QA~\cite{yu2019activitynet} introduced high-quality datasets focusing on actions within long videos. EgoSchema proposed the concept of temporal certificates to measure the inherent difficulty of video-related tasks. LongVideoBench~\cite{wu2024longvideobench} designed reasoning tasks, such as referring reasoning, to test models' ability to handle long videos, with a primary focus on segment retrieval. However, like other benchmarks such as VideoMME~\cite{fu2024video} and CinePine~\cite{rawal2024cinepile}, many of their question relies heavily on audio modality. In contrast, our work emphasizes understanding long videos purely in the visual modality. Additionally, works like LongVA~\cite{zhang2024long}, MLVU~\cite{zhou2024mlvu}, and LongVila~\cite{chen2024longvila} introduced the Visual Needle-In-A-Haystack (V-NIAH) experiment to evaluate models' ability to locate and retrieve visual information within long contexts. However, these experiments insert probes after sampling, ignoring the sampling challenges inherent to long video analysis, a critical issue for long-video processing. Our LSDBench specifically addresses the sampling problem by designing tasks on long videos with high necessary sampling density, aiming to evaluate the effectiveness of sampling strategies employed by video VLMs.

\minisection{Long-form Video Understanding}
In the domain of long-form video understanding, end-to-end models such as Qwen2VL~\cite{bai2025qwen2}, Qwen2.5VL~\cite{bai2025qwen2}, LongVila~\cite{zhang2024long}, LLaVA-OV~\cite{li2024llava}, and Gemini~\cite{team2024gemini} have demonstrated remarkable performance. However, these approaches still encounter notable challenges when it comes to sampling strategies. For instance, Qwen2.5VL introduced a dynamic sampling strategy aimed at balancing efficiency and performance, yet this approach operates at the global video level, overlooking redundancy in video segments. To adress this, some recent works try to reduce the redundancy from a token level. To tackle this issue, some recent works~\cite{bolya2022token,li2020tea,lin2019tsm,zhong2024lyra,yang2024visionzip,liu2025hybrid} attempt to reduce redundancy at the token level by leveraging visual token compression strategies. Additionally, some other methods~\cite{zhang2024omagent,arefeen2024vita,luo2024video} leverage the Retrieval-Augmented Generation (RAG)~\cite{lewis2020retrieval,jiang2023active} paradigm by segmenting long videos into smaller chunks, pre-processing them, and building an index to facilitate the quick retrieval of question-relevant details during inference, thus enabling accurate responses. While this approach effectively avoids the trade-off between sampling density and inference speed, it inherently makes RAG-based methods unsuitable for real-time scenarios.

\section{Dataset}
\label{sec:dataset}

\begin{figure*}[tb!] 
    \centering
    \includegraphics[width=\textwidth]{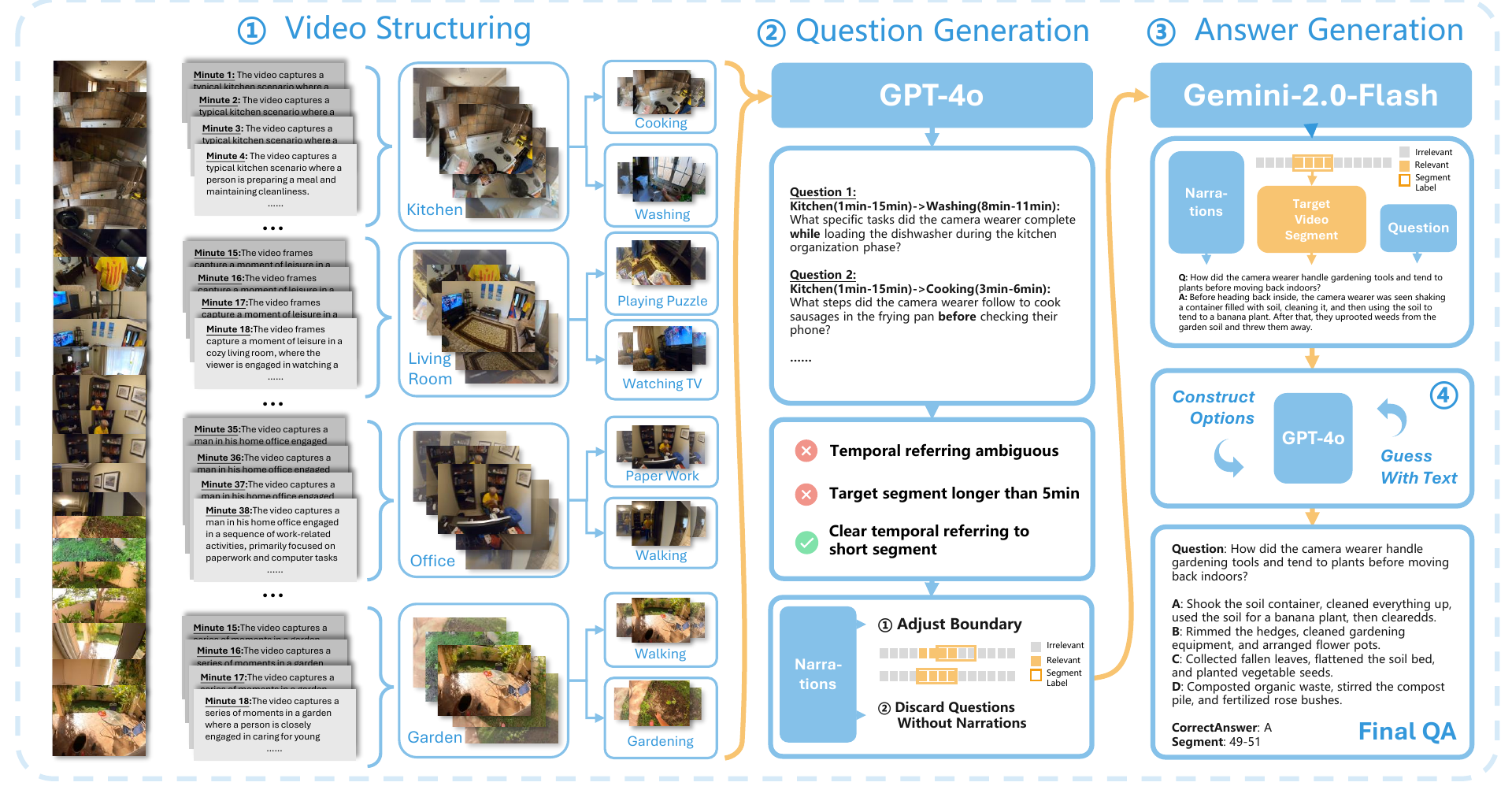} 
    \caption{\textbf{Pipeline of our four-stage data annotation pipeline.} We first segment the video into one-minute intervals and generate captions for each segment. These captions are then hierarchically clustered: the first layer clusters by scenes, while the second layer clusters by actions or events. Once the hierarchical tree structure is established, we utilize GPT-4o to generate questions, which are subsequently filtered and optimized through a two-round refinement process. For generating free-form answers to the questions, we leverage Gemini 2.0 Flash combined with narration-based annotations as auxiliary input. Finally, we construct multiple-choice options iteratively in an adversarial manner using GPT-4o.}
    \label{fig:data_pipeline}
\end{figure*}

In this section, we first introduce our definition of necessary sampling density (NSD) and explain how this concept serves as the foundation for designing our task.
Next, we detail the automated pipeline we developed for question-answer generation. Finally, we provide statistical information about the LSDBench dataset.

\subsection{Necessary Sampling Density}

We define the \textbf{necessary sampling density (NSD)} of a question as the minimum sampling density required, under uniform sampling,
to extract sufficient information from a video to answer a question, assuming no prior knowledge of the video.

Consider the question shown in \Cref{fig:teaser}, "What is the journey of the camera wearer in the video?" To answer this question, we need to watch the entire video, identify all locations visited, and determine their chronological order. Although the video is nearly an hour long and the task seems challenging, the camera wearer usually stays at each location for several minutes to tens of minutes, making the process more manageable. As a result, even if we employ a sparse sampling strategy—capturing only a single frame every few minutes—it is highly likely that a sampled frame will contain sufficient visual cues to identify the location. This suggests that even with such sparse sampling density, there is a significant chance of gathering enough visual cues to answer the question.

Furthermore, when the question is presented in a multiple-choice format and the distractors are not well-designed, the model can often guess the correct answer by leveraging the provided options, even if some critical cues are missed. The idea that "sampling just one frame per minute can allow us to answer a question about an hour-long video" may seem counterintuitive and unreliable. However, this type of low NSD problem fails to truly reveal the challenges of designing sampling strategies for long-video tasks since high NSD problem is more important and common.

\subsection{Task Design}

We aim to design a task for hour-long videos that has a sufficient NSD to emphasize the importance of sampling in long video tasks. Specifically, we construct questions about action sequences within short segments of long videos, such as Question 2 in \Cref{fig:teaser}. Since the intervals between actions and the duration of each action are only a few seconds, this ensures that the questions has a sufficiently high NSD. Meanwhile, temporal references are integrated into the question formulation to explicitly direct attention to the target segment, thereby reinforcing the necessity of understanding the video's overall temporal structure.
This design—featuring local segments with high NSD and a high proportion of redundant content across the video—makes our LSDBench more effective in evaluating the performance of VLMs in handling long videos and assessing the quality of their sampling strategies. 

\subsection{Annotation Pipeline}

We have meticulously designed a four-stage pipeline for fully automated QA generation. These four stages include (1) video hierarchical structuring, (2) question generation, (3) answer generation, and (4) multiple-choice question construction, as illustrated in \Cref{fig:data_pipeline}. We utilized the subset of long videos\footnote{The subset is filtered by HourVideo~\cite{chandrasegaran2024hourvideo}.} and corresponding action narrations annotated by human provided by EGO4D~\cite{grauman2022ego4d} for our data construction.

\minisection{Video Hierarchical Structuring}
Due to the extreme length of long videos, directly inputting the entire video and its annotations into an LLM for question generation is not feasible~\cite{chandrasegaran2024hourvideo}. Excessively long context inputs significantly degrade the instruction-following capabilities of LLMs and the quality of generated questions. Moreover, it becomes difficult to control the necessary sampling density and temporal certificate length for the questions.

To address this, we propose a hierarchical structuring of the video. First, we use the caption generated by GPT-4o for each individual minute of the video. These captions are then clustered hierarchically: high-level clustering is performed at the scene level, and low-level clustering is based on activities/events within each scene. A summary is simultaneously generated for each cluster.  This process produces a tree-structured temporal hierarchy of the video, facilitating targeted questioning for specific segments of the video.

\minisection{Question Generation}
We select nodes in the hierarchical structure with a duration of less than 5 minutes and use their summaries for question generation. We observed that directly providing narrations to LLM along with generating answers leads to reduced question diversity. To address this, we decouple question and answer generation into two independent stages. At this stage, LLM~\cite{hurst2024gpt,achiam2023gpt} generates questions solely based on the summaries of the clustered segments. This approach largely enhances the diversity of the questions. 

We also stipulate that the generated questions must include a referring expression that explicitly points to the relevant video segment to avoid ambiguity.

Next, we perform a two-round filtering process to refine the generated questions:

\begin{itemize}
    \item First round of filtering: We remove questions with unclear or ambiguous referring expressions. By providing LLM with only the question statement and the hierarchical structure, we validate whether the question clearly targets a specific segment. Questions leading to ambiguity or targeting segments too long are discarded.
    
    \item Second round of filtering: Since all previous stages rely on LLM-generated content, factual inaccuracies may arise. In this stage, we incorporate narrations from the corresponding time segment along with an additional 2 minutes of context on either side (before and after the segment) to verify the question. Questions totally unrelated to the narrations are removed, and the time boundaries of the target segment are adjusted based on the narrations.
\end{itemize}

\minisection{Answer Generation}
We use large multi-modal modal (LMM, here we use Gemini-2.0-Flash~\cite{team2024gemini}, which supports long video inputs) to generate answers.
The input to the model consists of all narrations within the target segment, the question, and corresponding video clip.
LMM organizes and generates the answer based on these inputs.

\minisection{Multiple-Choice Question Construction}
We leverage LLM to transform free-form answers into multiple-choice questions. To prevent scenarios where the correct answer can be easily guessed based on the question and options alone, we incorporate an adversarial option optimization process. Specifically, we simulate the process of guessing the correct answer by reasoning solely based on the question and options, without any visual information. The reasoning process is collected and used to iteratively refine the design of the options.

\subsection{Dataset Statistics}

\begin{figure}[tb!]
    \centering
    \begin{subfigure}{0.22\textwidth}
        \centering
        \adjustbox{valign=t}{\includegraphics[width=0.93\textwidth]{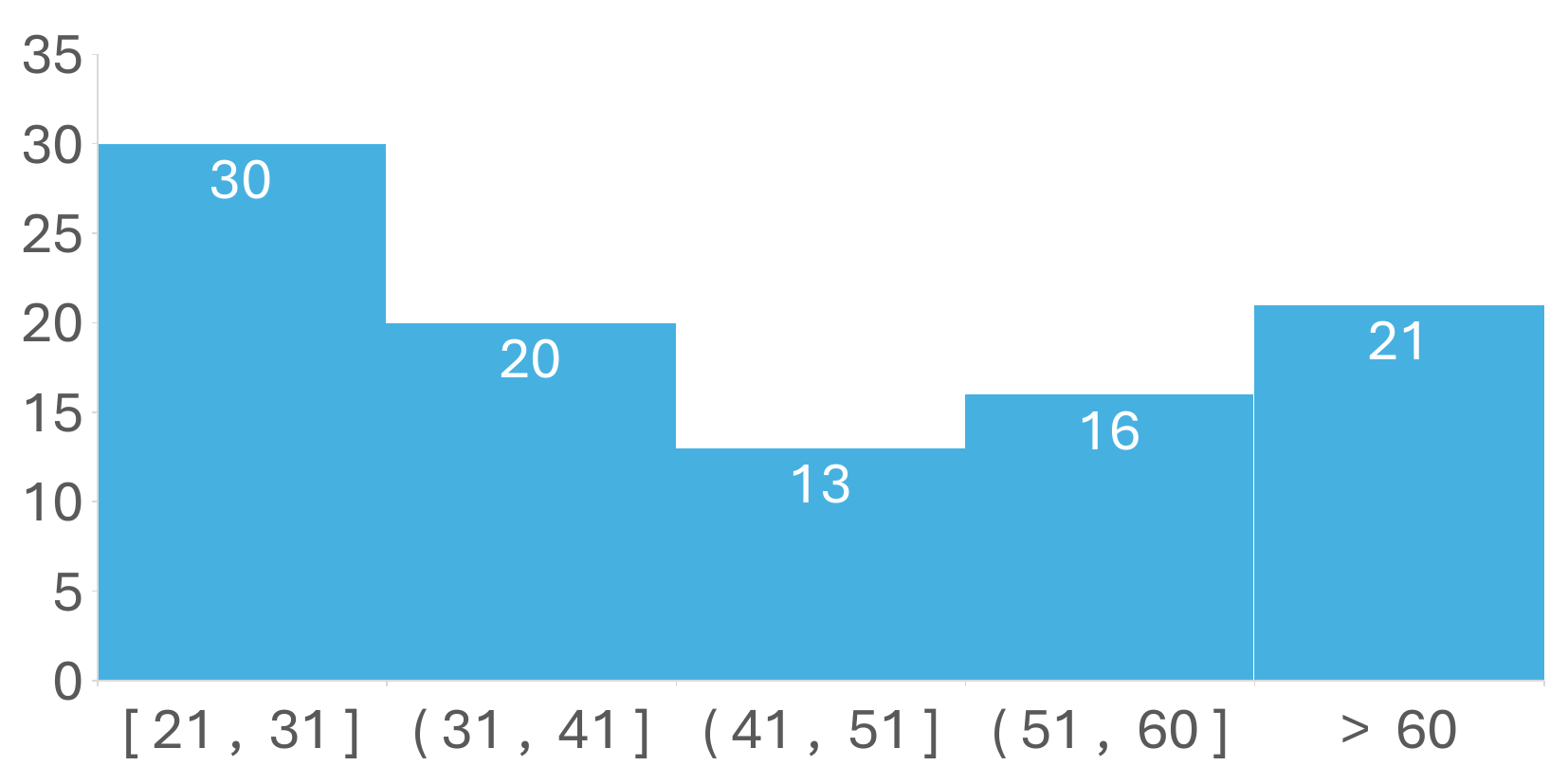} }
        \caption{Option length (words)}
        \label{fig:stat_subfig1}
    \end{subfigure}
    \begin{subfigure}{0.22\textwidth}
        \centering
        \adjustbox{valign=t}{\includegraphics[width=0.85\textwidth]{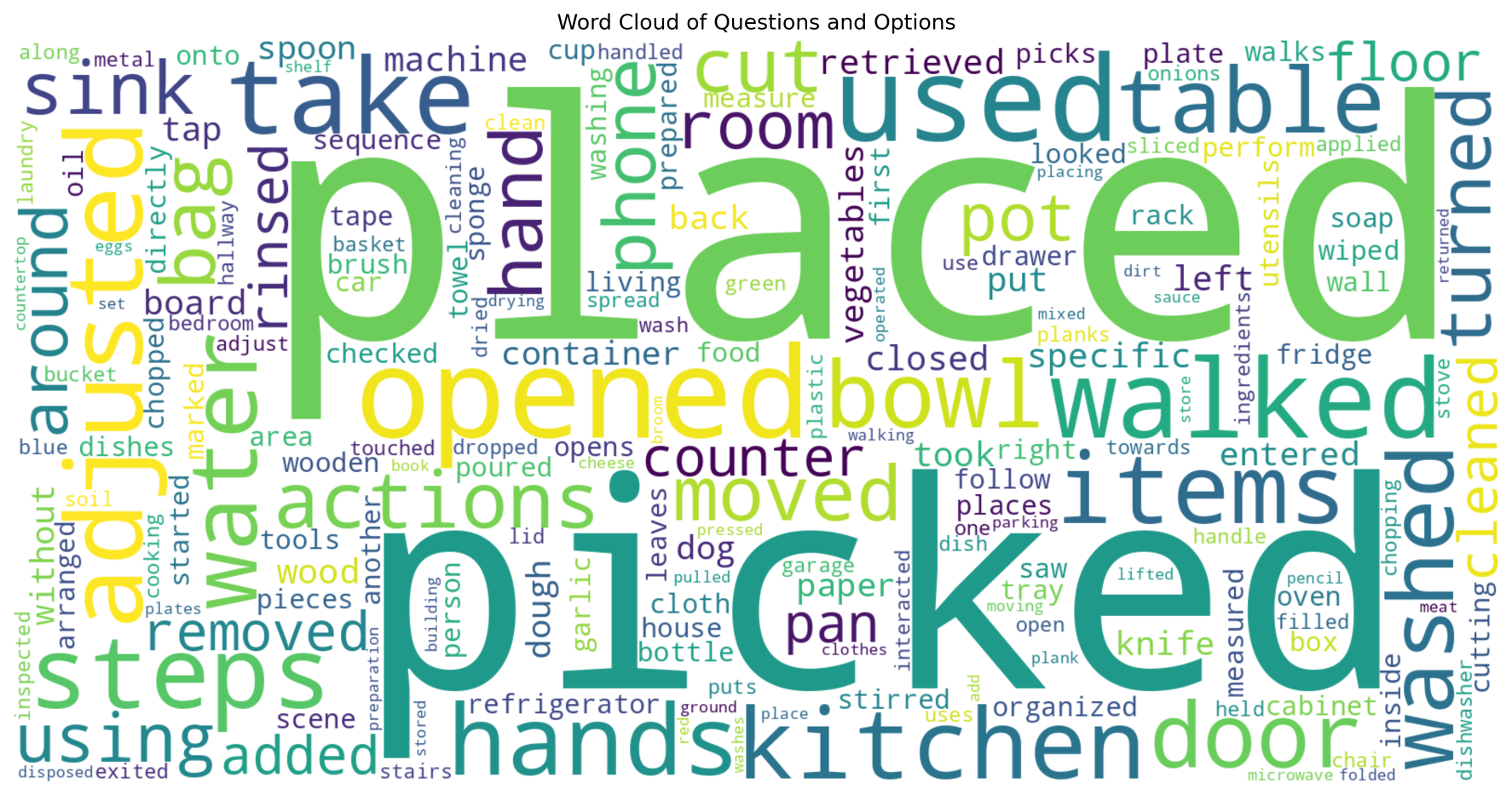}}
        \caption{Word cloud}
        \label{fig:stat_subfig2}
    \end{subfigure}
    
    \begin{subfigure}{0.22\textwidth}
        \centering
        \hspace{-0.5cm}
        \adjustbox{valign=t}{\includegraphics[width=0.95\textwidth]{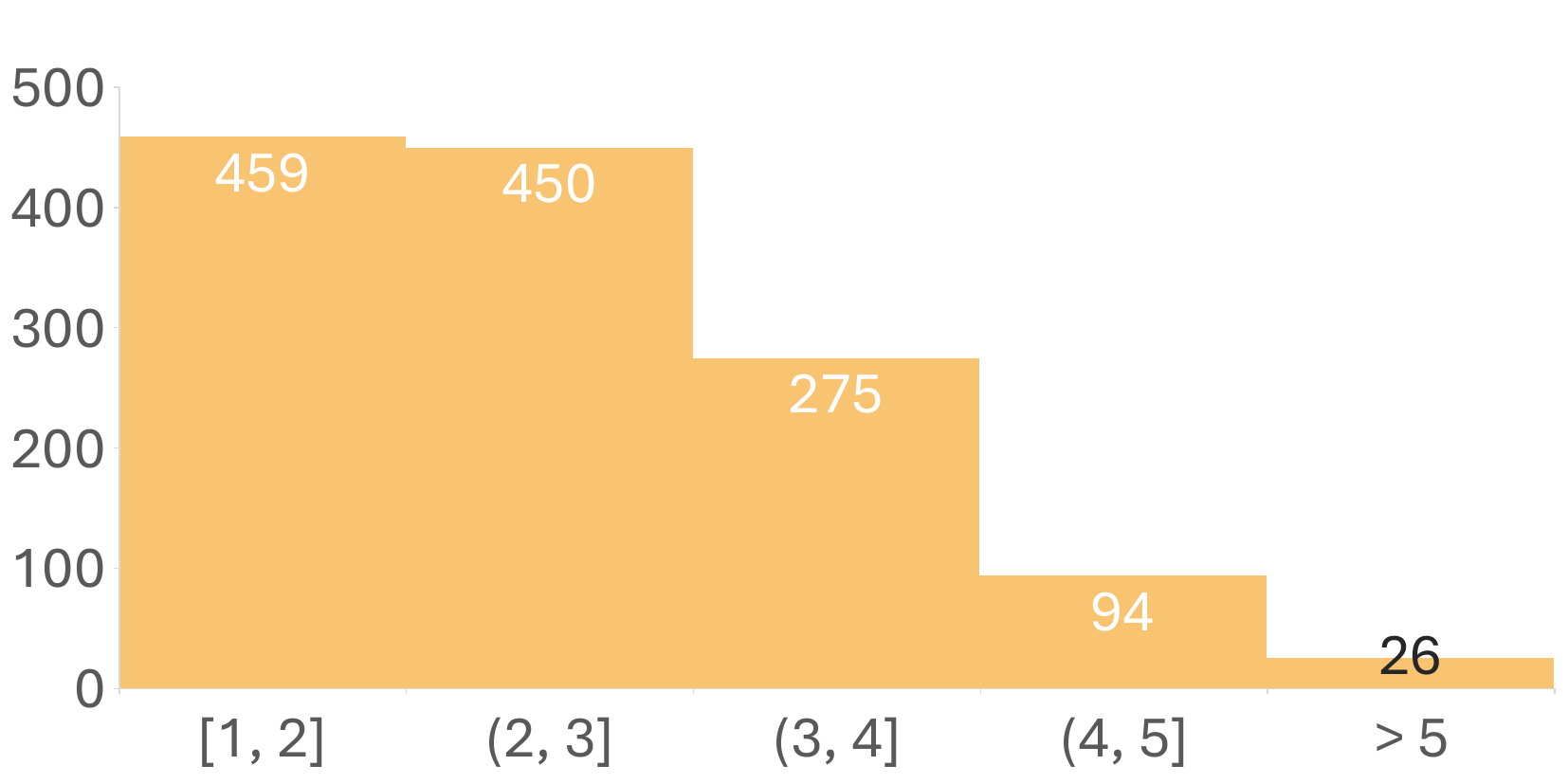}}
        \caption{Target segment duration (min)}
        \label{fig:stat_subfig3}
    \end{subfigure}
    \begin{subfigure}{0.2\textwidth}
        \centering
        \adjustbox{valign=t}{\includegraphics[width=\textwidth]{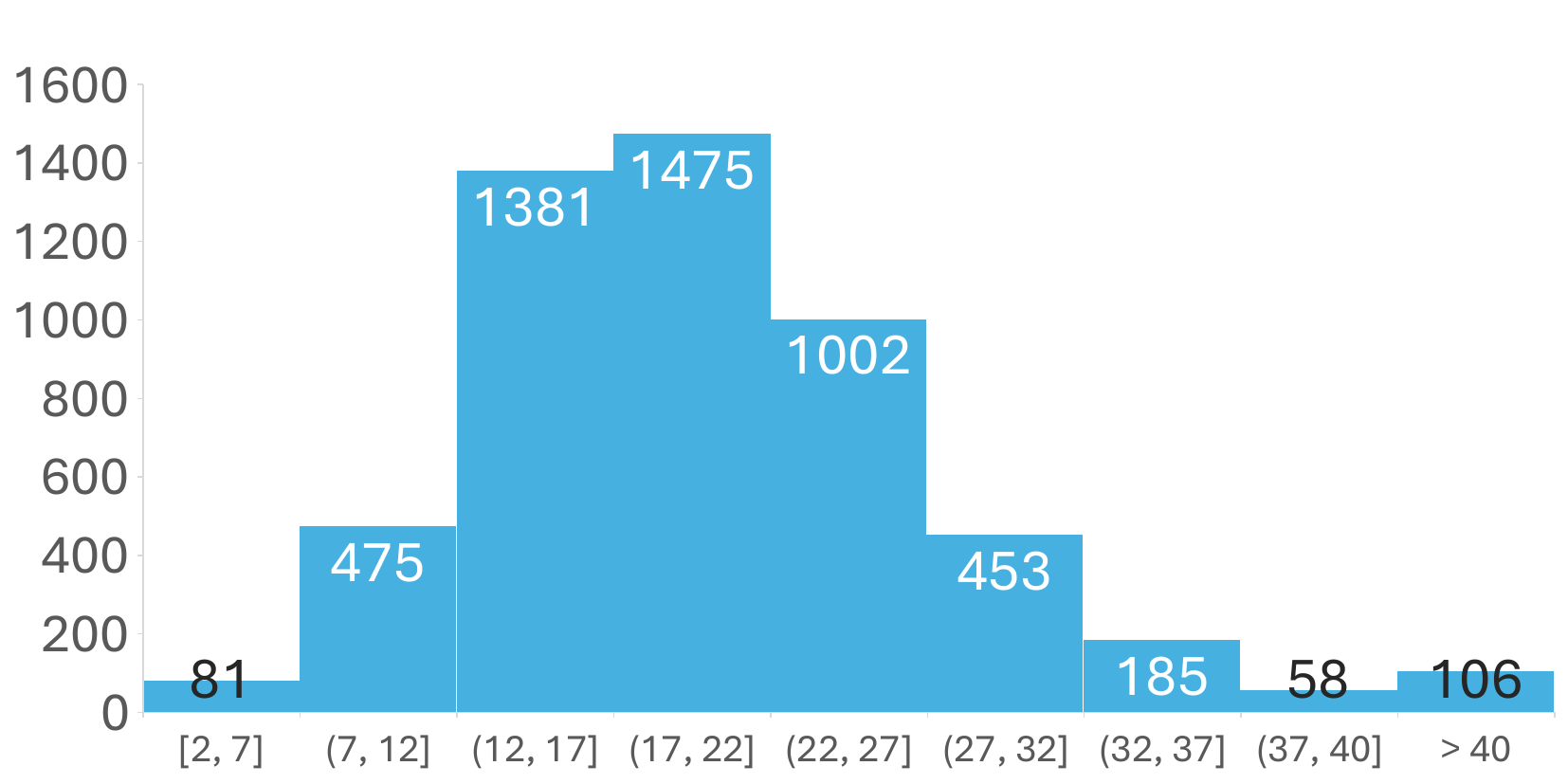}}
        \caption{Video duration (min)}
        \label{fig:stat_subfig4}
    \end{subfigure}

    \caption{Dataset statistics visualization.}
    \label{fig:mainfig}
\end{figure}

The dataset comprises 1304 QA pairs derived from 400 videos, with an average option length of 20.7 words.
The target segments have an average duration of 3 minutes, with the majority falling within the 2--4 minute range, and very few exceeding 5 minutes.
The analyzed videos have an average duration of 45.39 minutes, with lengths ranging from 20.32 to 115.32 minutes.
A QA generation pipeline was applied to 400 videos, initially producing 11,342 questions.
However, through a rigorous multi-stage filtering process, this number was reduced to 1304 finalized QA pairs,
representing a retention rate of 11.6\%, ensuring high data quality.

\section{Method}
\label{sec:method}

\begin{figure}[tb!]
    \centering
    \includegraphics[width=\linewidth]{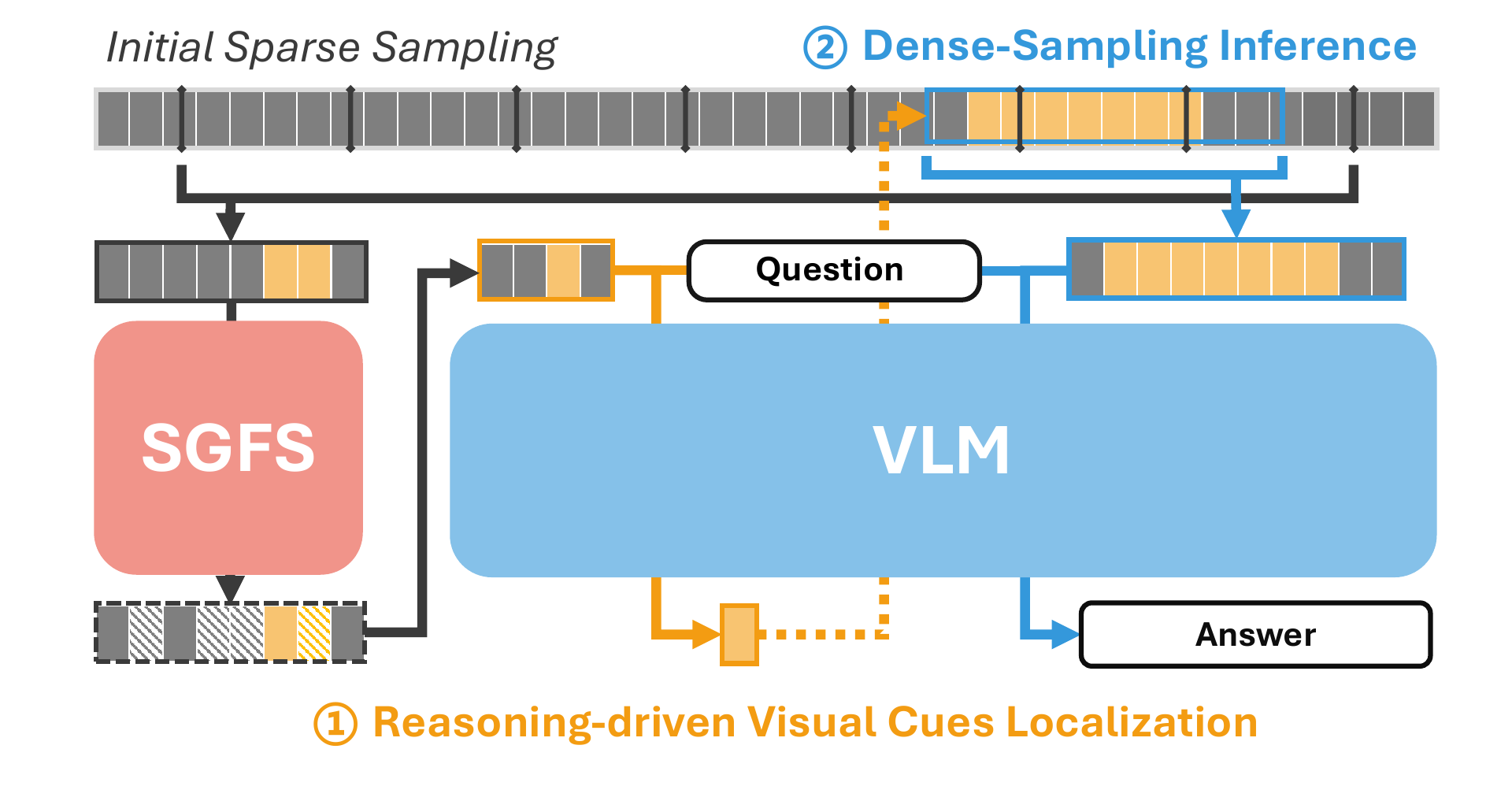}
    \caption{Overview of our Reasoning-Driven Hierarchical Sampling (RHS) framework.}
    \label{fig:method}
\end{figure}

\subsection{Reasoning-Driven Hierarchical Sampling}

To address the challenge of high necessary sampling density in long videos efficiently,
we propose \textbf{Reasoning-Driven Hierarchical Sampling (RHS)}, a two-stage framework tailored for long video tasks. 

Our core idea is to leverage large model's reasoning capability to exploit the temporal structure of the video in a top-down manner.
Specifically, we aim to infer the approximate locations of short-term temporal information by reasoning over the sparse sampled keyframes. For example, we can reason about event progression, temporal order, or scene transitions. Since short-term temporal information is largely derived from motion, and adjacent frames tend to be spatially similar, we can identify regions where dense sampling is necessary by reasoning over the spatial semantics of sparsely sampled keyframes. This demand-driven sampling approach not only significantly reduces the computational overhead caused by processing redundant segments but also preserves the ability to capture highly fine-grained temporal details.

As shown in \Cref{fig:method}, our method comprises two stages:
\wcircled{1} {Reasoning-driven Visual Cues Localization}, 
and \wcircled{2} {Dense-Sampling Inference}.
In the first stage, we perform Sparse Sampling to select keyframes. The video is then partitioned into segments, with midpoints between keyframes as boundaries. These keyframes and their indices are fed into VLM, which reasons over the question and spatio-temporal context (e.g., chronological order, causal logic, co-occurrence of objects) to identify segments that has visual cues relevant to the question. In the second stage, the selected segments are densely sampled to gather finer-grained temporal details. The sampled frames are concatenated and input into the VLM for question answering. 

This framework focuses dense sampling only on key segments, significantly reducing redundant inputs while handling high NSD tasks. However, sparse sampling in first stage may occasionally miss critical information, motivating us to explore more effective sampling strategies.

\subsection{Semantic-Guided Frame Selector}

Our primary objective is to ensure that the sparsely sampled keyframe set captures as much essential information as possible. Each keyframe should effectively represent a distinct temporal segment of the video, facilitating more accurate reasoning by the VLM during the first stage.

To achieve this, we propose the efficient, plug-and-play \textbf{Semantic-Guided Frame Selector (SGFS)} that selects a subset of frames containing the richest information from the initial larger frame set.

Specifically, we begin by uniformly sampling $n$ frames without any prior knowledge. Our objective is to select a subset of keyframes with target size $k$ such that the sum of similarities between any two adjacent frames in the subset is minimized, thereby maximizing the diversity of the selected frames and the overall information captured. Using a lightweight visual encoder (e.g. SigLIP2~\cite{tschannen2025siglip,radford2021learning}), we extract embeddings for these frames. Formally, for a frame $F_i$, we compute its embedding vector $e_i$ as: $e_i = \text{Encoder}(F_i), \quad i \in \{1, 2, \ldots, n\}$,
where $e_i \in \mathbb{R}^d$ and $d$ is the dimension of the embedding space.
Next, we compute pairwise cosine similarity between frames to form a similarity matrix $S$. For two frames with indices $i$ and $j$, the similarity is given by:
\begin{equation}
S_{ij} = \frac{e_i \cdot e_j}{||e_i|| \cdot ||e_j||}.
\end{equation}

To improve robustness and prevent the selected subset from being overly concentrated in some period, we introduce distance penalty for the temporal proximity of adjacent frames. For two frames with indices $i$ and $j$:
\begin{equation}
P_{ij} = -\lambda * \left|\frac{i}{n}-\frac{j}{n}\right|^\beta,
\end{equation}
where $P_{ij}$ denotes the distance penalty between these two frames, $\lambda$ and $\beta$ are hyperparameters.
This results in a final weight matrix $W$ where:
\begin{equation}
W_{ij} = S_{ij} + P_{ij}.
\end{equation}

Formally, the optimal subset of keyframes can be selected by minimizing the sum of pairwise weights between consecutive frames in the subset. Let the selected set of keyframe indices be denoted as $\mathcal{K} = \{K_1, K_2, \ldots, K_k\}$, where $K_1 < K_2 < \dots < K_k$. The selection objective can be written as:
\begin{equation}
\mathcal{K} = \mathop{\arg\min}_{\mathcal{K} \subseteq \{1, 2, \ldots, n\}, |\mathcal{K}| = k} \sum_{i=1}^{k-1} W_{K_i, K_{i+1}}.
\end{equation}
We use a dynamic programming algorithm as shown in Algorithm~\ref{alg:dp_select} to compute the optimal subset of keyframes efficiently.

\begin{algorithm}
\caption{Frame Sampling Algorithm}
\label{alg:dp_select}
\SetAlgoLined
\SetAlgoNoEnd
\small
\KwIn{$n$ frames, target count $k$, weight matrix $W$}
\KwOut{Selected frame indices}

$dp[0..n, 0..k] \gets \infty$, $trace[0..n, 0..k] \gets -1$, $dp[0,0] \gets 0$\;

\For{$j = 1$ \KwTo $k$}{
    \For{$i = j$ \KwTo $n$}{
        \For{$p = j-1$ \KwTo $i-1$}{
            \If{$dp[p,j-1] + W[p,i] < dp[i,j]$}{
                $dp[i,j] \gets dp[p,j-1] + W[p,i]$\;
                $trace[i,j] \gets p$\;
            }
        }
    }
}

$sel \gets \emptyset, i \gets n, j \gets k$\;
\While{$j > 0$}{
    $sel \gets sel \cup \{i\}, i \gets trace[i,j], j \gets j-1$\;
}
\Return{$reverse(sel)$}\;
\end{algorithm}

\section{Experiments}
\label{sec:experiment}

\begin{table}[t!]
\centering
\resizebox{\columnwidth}{!}{ 
\renewcommand\arraystretch{0.95} 
    \begin{tabular}{l|c|c|c|c}
    \toprule
    \textbf{Method} & \textbf{Sampling} & \textbf{Frames} & \textbf{SD (f/s)} & \textbf{Acc (\%)} \\
    \midrule
    \multicolumn{5}{c}{\textbf{\textit{Only Text}}} \\
    \midrule
    Gemini-2.0-Flash & - & - & - & 29.1 \\
    Qwen2-VL & - & - & - & 30.5 \\
    LongVila & - & - & - & 36.9 \\
    Qwen2.5-VL & - & - & - & 30.2 \\
    InternVideo2.5 & - & - & - & 36.1 \\
    \midrule

    \multicolumn{5}{c}{\textbf{\textit{Oracle} (avarage 3 min)}} \\
    \midrule
    Gemini-2.0-Flash \textit{Oracle} & 1-FPS & 180 & 1 & 64.8 \\
    \midrule
    \multirow{3}{*}{LongVila \textit{Oracle}} & \multirow{3}{*}{Fixed} & 64 & 0.35 & 54.2 \\
     &  & 128 & 0.7 & 52.0 \\
     &  & 256 & 1.4 & 51.9 \\
    \midrule
    \multirow{3}{*}{} & \multirow{3}{*}{} & 64 & 0.35 & 58.3 \\
    InternVideo2.5 \textit{Oracle} & Fixed & 128 & 0.7 & 58.9 \\
      &  & 256 & 1.4 & 59.1 \\
    \midrule
    \multirow{5}{*}{Qwen2.5-VL \textit{Oracle}} & 0.1-FPS & 18 & 0.1 & 49.6 \\
     & 0.2-FPS & 36 & 0.2 & 52.1 \\
     & 0.25-FPS & 45 & 0.25 & 53.2 \\
     & 0.5-FPS & 90 & 0.5 & 54.9 \\
     & 1-FPS & 180 & 1 & 56.4 \\
    \midrule

    \multicolumn{5}{c}{\textbf{\textit{Full Video (average 45 min)}}} \\
    \midrule

    \multirow{3}{*}{LongVA} & \multirow{3}{*}{Fixed} & 256 & 0.1 & 31.3 \\
     &  & 512 & 0.2 & 33.0 \\
     &  & 1024 & 0.4 & 32.5 \\
    \midrule
     
    Qwen2-VL & Fixed & 256 & 0.1 & 48.0 \\
    \midrule
    \multirow{2}{*}{Qwen2.5-VL} & \multirow{2}{*}{Fixed} & 256 & 0.1 & 50.1 \\
     &  & 768 & 0.3 & 52.5 \\
    \midrule
    LongVila & Fixed & 256 & 0.1 & 49.8 \\
    \midrule
    InternVideo2.5 & Fixed & 256 & 0.1 & 50.1 \\
    \midrule
    \rowcolor{gray!20} Qwen2.5-VL (Ours) & RHS & 225 & 1 & 52.2 \\
    \midrule
    \textit{\textcolor{gray}{Gemini-2.0-Flash}} & \textit{\textcolor{gray}{1-FPS}} & \textit{\textcolor{gray}{2700}} & \textit{\textcolor{gray}{1.00}} & \textit{\textcolor{gray}{56.2}} \\
    \bottomrule
    \end{tabular}

    }
\caption{\textbf{Performance comparison of different models and sampling strategies.} We present three testing settings in total: Only Text, Oracle, and Full Video. In the Only Text setting, the model is provided with no visual information whatsoever. The Oracle setting involves using the annotated target segment as the video input, while the Full Video setting provides the complete long video as input. The "Sampling" column lists the sampling strategies used: FPS represents sampling at fixed time intervals, fixed denotes uniform sampling with a fixed number of frames, and 2stage refers to the method we propose. Under each sampling strategy, the average number of sampled frames during evaluation on the LSDBench dataset is reported in the "Frames" column, along with the corresponding sampling density (SD). }
\label{tab:model_comparison}
\end{table}

In this section, we present the results of state-of-the-art long-video VLMs, as well as our proposed 2-stage method on our benchmark.
We also conduct ablation studies on RHS and the SGFS.

\subsection{Experiment Setting}

To comprehensively evaluate the performance of long-video VLMs on our task, we selected the following models: Gemini-2.0-Flash~\cite{team2024gemini}, Qwen-2.5VL~\cite{bai2025qwen2}, Qwen-2VL~\cite{bai2025qwen2}, LongVA~\cite{zhang2024long}, LongVila~\cite{chen2024longvila}, InternVideo2.5~\cite{chen2024internvl}

We experimented with different sampling strategies and frame count settings for these models to explore the relationship between sampling configurations and model performance on the task. The specific configurations are detailed in the table provided. It is worth noting that under the 1-FPS sampling density setting, the number of sampled frames exceeded the context length limitation of most models. Therefore, for models that could not support a sampling rate of 1-FPS, we adopted a fixed sampling frame count setting for evaluation. To ensure fairness, all other generation configurations followed their original settings.

Additionally, we report the performance of our 2-stage method applied on Qwen2.5-VL. In our approach, we use Siglip2~\cite{tschannen2025siglip} as the encoder for the sampling module. The initial sampling rate is set to 4 frames per minute, and the frame retention ratio for the sampling module is set to 0.25. The length penalty $\lambda$ is 10.0 and $\beta$ is 0.3.

\begin{figure}[tb!]
    \centering
    \includegraphics[width=.98\linewidth]{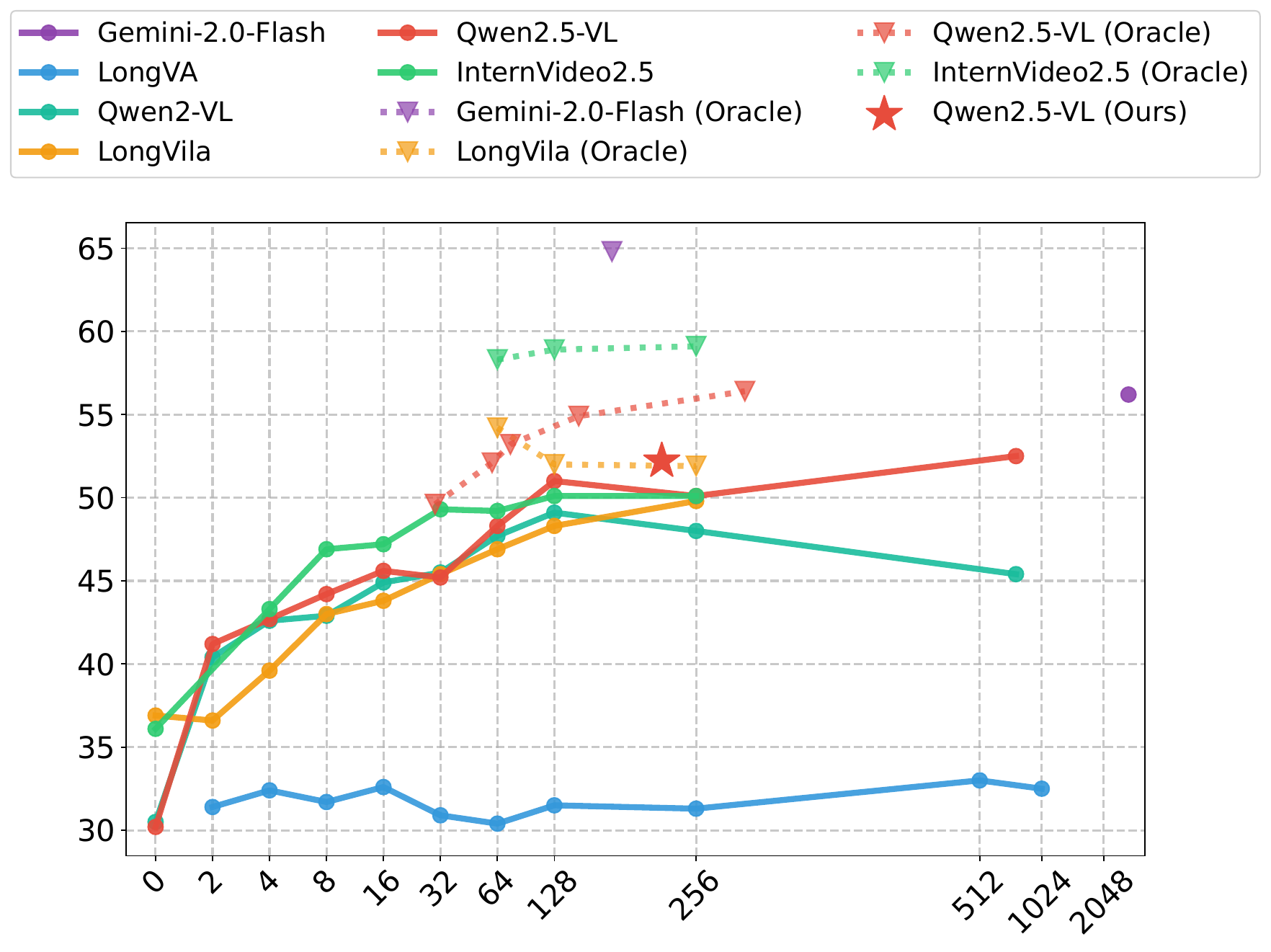} 
    \label{fig:acc_frms}
    \caption{
        \textbf{The line graph illustrates the relationship between the number of sampled frames (x-axis) and accuracy on LSDBench (y-axis).}
        Solid lines represent results under the Full Video setting, while dashed lines with inverted triangles correspond to the Oracle setting.
        The gap between the \textit{Oracle} and global uniform sampling highlights the potential for improved sampling strategies in long-video VLMs.
    }
\end{figure}

\subsection{Results On Benchmark}

We conducted the \textit{Oracle} experiment by exclusively using the annotated target segments of the videos as input. This setup was designed to evaluate the upper-bound performance achievable under ideal sampling conditions, where sufficient visual information is guaranteed, given a fixed total number of sampled frames. Comparatively, the results obtained from global uniform sampling of the full video input reveal a striking contrast. Even with several times more sampled frames under uniform sampling, the performance remains significantly below that of the \textit{Oracle} setup.

As illustrated in \Cref{tab:model_comparison}, the benefits of increasing the number of sampled frames diminish progressively under global uniform sampling. For example, when increasing the sampled frames from 128 to 256, many models even exhibit a decline in accuracy. Additionally, further increasing the sampling density often results in input lengths exceeding the maximum context length supported by the models. While Gemini-2.0-Flash is capable of handling an ultra-long context length of up to 1 million tokens, even at a 1 fps sampling rate, its global uniform sampling results lag by 8.6\% compared to the \textit{Oracle} setup. These findings underscore the inefficiency of global uniform sampling in utilizing the sampled frames. The redundant inputs not only fail to enhance performance but also introduce noise, negatively impacting the model's ability to realize its full potential. The gap between \textit{Oracle} performance and global uniform sampling highlights a substantial opportunity for improvement in sampling strategies for long-video VLM.

Our proposed RHS demonstrates significant efficiency gains. For instance, on Qwen2.5-VL, it achieves an accuracy of 52.2\% using just 225 sampled frames, which is comparable to the global uniform sampling result achieved with the model's maximum setting of 768 frames. Moreover, it surpasses the result obtained with 256 globally uniformly sampled frames by 2.1\%. Notably, our method achieves these results without requiring any additional training of the model. This demonstrates that improving the sampling strategy can lead to better performance with substantially fewer sampled frames, validating the effectiveness of our approach.

\begin{table}[t!]
\centering
\resizebox{\columnwidth}{!}{ 
\renewcommand\arraystretch{0.95} 
    \begin{tabular}{l|c|c|l}
    \toprule
    \multicolumn{2}{c|}{\textbf{Sampling}} & \textbf{Frames} & \textbf{Acc (\%)} \\
    \midrule
    \multicolumn{4}{c}{\textbf{1 Stage Methods}} \\
    \midrule
    \multicolumn{2}{c|}{Fixed 32} & 32 & 45.2 \\
    \multicolumn{2}{c|}{Fixed 128$\rightarrow$32} & 32 & 48.3 (+3.1) \\
    \midrule
    \multicolumn{2}{c|}{Fixed 64} & 64 & 48.3 \\
    \multicolumn{2}{c|}{Fixed 128$\rightarrow$64} & 64 & 47.3 (-1.0) \\
    \multicolumn{2}{c|}{Fixed 256$\rightarrow$64} & 64 & 48.7 (+0.4) \\
    \midrule
    \multicolumn{2}{c|}{Fixed 128} & 128 & 51 \\
    \multicolumn{2}{c|}{Fixed 256$\rightarrow$128} & 128 & 51.3 (+0.3) \\
    \multicolumn{2}{c|}{Fixed 512$\rightarrow$128} & 128 & 51.8 (+0.8) \\
    \midrule
    \multicolumn{2}{c|}{Fixed 256} & 256 & 50.1 \\
    \multicolumn{2}{c|}{Fixed 512$\rightarrow$256} & 256 & 49.6 (-0.5) \\
    \multicolumn{2}{c|}{Fixed 1024$\rightarrow$256} & 256 & 51.0 (+0.9) \\
    \midrule
    \multicolumn{2}{c|} {Fixed 768} & 768 & 52.5 \\
    \midrule
    \multicolumn{2}{c|}{0.1-FPS} & 262 & 50.2 \\
    \multicolumn{2}{c|}{0.2-FPS} & 523 & 51.6 \\
    \multicolumn{2}{c|}{0.25-FPS} & Out of context length & - \\
    \multicolumn{2}{c|}{0.3-FPS} & Out of context length & - \\
    \midrule
    \multicolumn{4}{c}{\textbf{2 Stage Methods}} \\
    \midrule
    \textbf{Sparse Sampling} & \textbf{Dense Sampling} & \textbf{Frames} & \textbf{Acc (\%)} \\
    \midrule
    \textit{Oracle} & \textit{0.25-FPS} & \textit{45} & \textit{53.2} \\
    1-FPM & 0.25-FPS & 90 & 49.4 \\
    \rowcolor{gray!20} 4-FPM$\rightarrow$1-FPM & 0.25-FPS & 90 & 49.9 (+0.5) \\
    \midrule
    \textit{Oracle} & \textit{0.5-FPS} & \textit{90} & \textit{54.9} \\
    1-FPM & 0.5-FPS & 135 & 49.4 \\
    \rowcolor{gray!20} 4-FPM$\rightarrow$1-FPM & 0.5-FPS & 135 & 50.2 (+0.8) \\
    \midrule
    \textit{Oracle} & \textit{1-FPS} & \textit{180} & \textit{56.4} \\
    1-FPM & 1-FPS & 225 & 51.0 \\
    \rowcolor{gray!20} 4-FPM$\rightarrow$1-FPM & 1-FPS & 225 & 52.2 (+1.2) \\
    \bottomrule
    \end{tabular}
    }
\caption{\textbf{Comparison with different sampling strategies and ablation studies on SGFS.} We conducted experiments on Qwen2.5-VL. The "Frames" column shows the average total sampled frames for different settings on LSDBench. In the 1-stage section, we compare global uniform sampling with direct SGFS results. In the 2-stage section, we present results for the Oracle and RHS methods. FPM denotes sampling one frame per minute. The arrow (→) indicates the initial uniform sample count on the left and the retained frames after SGFS filtering, along with the average sampling density, on the right.}
\label{tab:model_sampling_new}
\end{table}

\subsection{Ablation Study}

As shown in the \Cref{tab:model_sampling_new}, we conducted an ablation study to evaluate our proposed method. For the experiments, we selected Qwen2.5-VL as the VLM. Initially, we tested global sampling approach, where we fixed the number of sampled frames passed into the VLM. When the initial sampled frames reached 4 times the target frame count, our sampling module was able to improve the accuracy of the VLM to varying extents, even though our module only considers semantic information.
We provide additional visualizations of the frames sampled with using SGFS in Appendix for reference. Notably, our sampling module is lightweight, when compared to the computational cost of the VLM processing the video, the overhead introduced by the sampling process is almost negligible. 

When employing the complete RHS method, where the sampling module is used during the initial sparse sampling phase, the results demonstrate significant improvements over direct uniform sampling. For instance, in the first stage, we set the initial sampling rate to 4 frames per minute, with the sampling module retaining 1 in 4 frames. In the second stage, we adopted a 1-FPS sampling rate. This configuration achieved an accuracy of 52.2\%, with the total number of sampled frames averaging around 225. In contrast, using a global uniform sampling method at 0.1 FPS resulted in an accuracy of only 50.23\%, despite averaging 263 sampled frames. These findings highlight the efficiency advantage of our method in terms of sampling density.

\begin{figure}[tb!]
    \centering
    \begin{minipage}{0.22\textwidth}
        \centering
        \includegraphics[width=\textwidth]{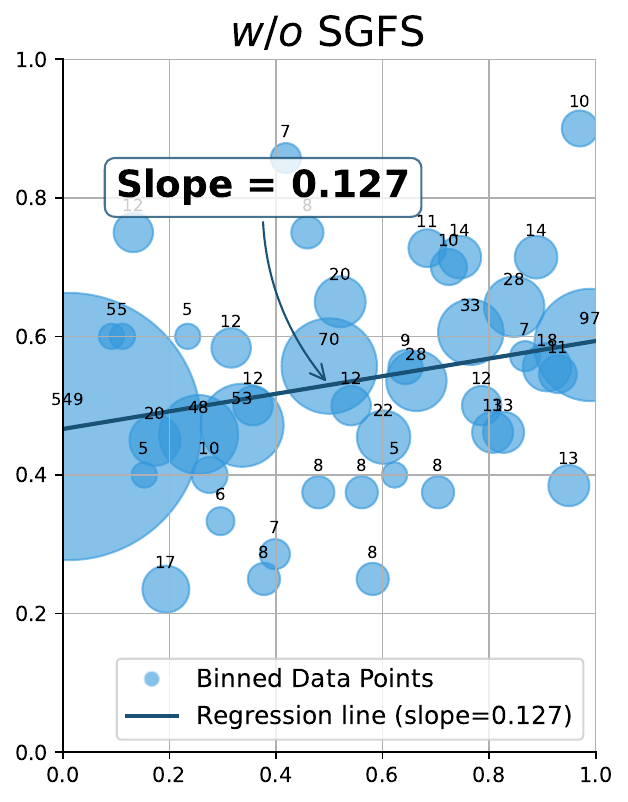}
        \label{fig:acc_cr_uni}
    \end{minipage}
    \begin{minipage}{0.22\textwidth}
        \centering
        \includegraphics[width=\textwidth]{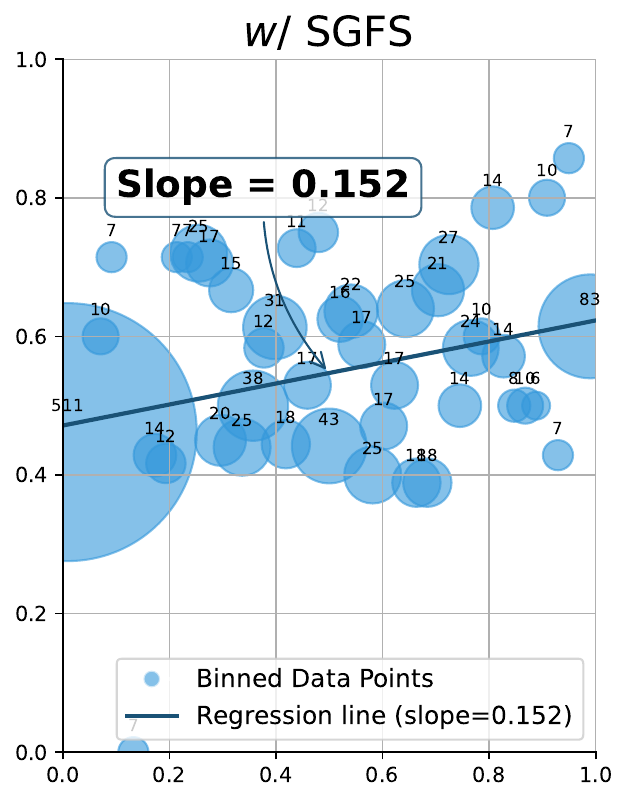}
        \label{fig:acc_cr_sgfs}
    \end{minipage}

    \caption{Correlation between overall accuracy (y-axis) and the coverage rate of the first-stage predictions on the ground truth target segments (x-axis).}
    \label{fig:acc_cr}
\end{figure}

Additionally, as illustrated in \Cref{fig:acc_cr}, we plotted the correlation between overall accuracy and the coverage rate of the first-stage predictions on the ground truth target segments. It is evident that with the inclusion of our SGFS, the correlation line exhibits a steeper slope compared to uniform sparse sampling. Furthermore, under the same coverage rate, our approach achieves higher accuracy. This highlights that the frame set selected by our sampling module retains more valuable information, allowing the VLM to better localize and reason about visual cues.
Further ablation studies on hyperparameters are provided in the Appendix.

\section{Conclusion}
\label{sec:conclusion}

In this work, we present LSDBench, the first benchmark designed to emphasize Necessary Sampling Density (NSD) in long-video tasks. LSDBench evaluates LVLMs in high-NSD scenarios with densely packed actions and temporally structured questions. To tackle the sampling challenge, we propose a Reasoning-Driven Hierarchical Sampling framework and a Semantic-Guided Frame Selector, improving efficiency and accuracy by focusing on key segments and informative frames. Our contributions set a new standard for advancing long-video understanding in high-NSD tasks.

\clearpage
{
    \small
    \bibliographystyle{ieeenat_fullname}

}

\clearpage
\setcounter{page}{1}
\maketitlesupplementary

\section{Additional Experiment Results}
\label{sec:add-exp}

We present in Table~\ref{tab:more_result} the accuracy of the model on LSDBench when the full video is used as input under varying numbers of sampled frames. Additionally, in Table~\ref{tab:sampling_ablation_2}, we provide the accuracy performance when using SGFS and RHS under different initial sampling frame counts and SGFS target frame settings.

\begin{table}[htbp]
\centering
\resizebox{\columnwidth}{!}{
\small
\begin{tabular}{lccc}
\toprule
\textbf{Model} & \textbf{Sampling} & \textbf{Frames} & \textbf{Accuracy} \\
\midrule
Gemini-2.0-Flash & 1-FPS & 2700 & 56.2 \\
Gemini-2.0-Flash Oracle & 1-FPS & 180 & 64.8 \\
Gemini-2.0-Flash & Text & 0 & 29.1 \\
\midrule
LongVA & Fixed & 2 & 31.4 \\
LongVA & Fixed & 4 & 32.4 \\
LongVA & Fixed & 8 & 31.7 \\
LongVA & Fixed & 16 & 32.6 \\
LongVA & Fixed & 32 & 30.9 \\
LongVA & Fixed & 64 & 30.4 \\
LongVA & Fixed & 128 & 31.5 \\
LongVA & Fixed & 256 & 31.3 \\
LongVA & Fixed & 512 & 33.0 \\
LongVA & Fixed & 1024 & 32.5 \\
\midrule
Qwen2-VL & text & 0 & 30.5 \\
Qwen2-VL & Fixed & 2 & 40.4 \\
Qwen2-VL & Fixed & 4 & 42.6 \\
Qwen2-VL & Fixed & 8 & 42.9 \\
Qwen2-VL & Fixed & 16 & 44.9 \\
Qwen2-VL & Fixed & 32 & 45.5 \\
Qwen2-VL & Fixed & 64 & 47.7 \\
Qwen2-VL & Fixed & 128 & 49.1 \\
Qwen2-VL & Fixed & 256 & 48.0 \\
Qwen2-VL & Fixed & 768 & 45.4 \\
\midrule
LongVila & Text & 0 & 36.9 \\
LongVila & Fixed & 2 & 36.6 \\
LongVila & Fixed & 4 & 39.6 \\
LongVila & Fixed & 8 & 43.0 \\
LongVila & Fixed & 16 & 43.8 \\
LongVila & Fixed & 32 & 45.4 \\
LongVila & Fixed & 64 & 46.9 \\
LongVila & Fixed & 128 & 48.3 \\
LongVila & Fixed & 256 & 49.8 \\
\midrule
Qwen2.5-VL & Only-Text & 0 & 30.2 \\
Qwen2.5-VL & Fixed & 2 & 41.2 \\
Qwen2.5-VL & Fixed & 4 & 42.7 \\
Qwen2.5-VL & Fixed & 8 & 44.2 \\
Qwen2.5-VL & Fixed & 16 & 45.6 \\
Qwen2.5-VL & Fixed & 32 & 45.2 \\
Qwen2.5-VL & Fixed & 64 & 48.3 \\
Qwen2.5-VL & Fixed & 128 & 51.0 \\
Qwen2.5-VL & Fixed & 256 & 50.1 \\
Qwen2.5-VL & Fixed & 768 & 52.5 \\
Qwen2.5-VL & 2-Stage & 45+180 & 52.2 \\
\midrule
InternVideo2.5 & Text & 0 & 36.1 \\
InternVideo2.5 & Text & 4 & 43.3 \\
InternVideo2.5 & Fixed & 4 & 43.3 \\
InternVideo2.5 & Fixed & 8 & 46.9 \\
InternVideo2.5 & Fixed & 16 & 47.2 \\
InternVideo2.5 & Fixed & 32 & 49.3 \\
InternVideo2.5 & Fixed & 64 & 49.2 \\
InternVideo2.5 & Fixed & 128 & 50.1 \\
InternVideo2.5 & Fixed & 256 & 50.1 \\
\bottomrule
\end{tabular}
}
\caption{Performance comparison of different models and sampling settings.}
\label{tab:more_result}
\end{table}

\begin{table}[htbp]
    \centering
    \begin{tabular}{cccc}
    \toprule
        \textbf{Initial Sampling} & \textbf{Acc (\%)} \\ 
        \midrule
        1-FPM   & 51.0  \\ 
        \midrule
        1-FPM $ \rightarrow$ 0.50-FPM   & 50.1  \\ 
        2-FPM $ \rightarrow$ 0.25-FPM & 48.7 \\ 
        2-FPM $ \rightarrow$ 0.50-FPM  & 50.1 \\ 
        2-FPM $ \rightarrow$ 1.00-FPM  & 51.5 \\ 
        4-FPM $ \rightarrow$ 0.50-FPM   & 51.8 \\ 
        \rowcolor{gray!20} 4-FPM $ \rightarrow$ 1.00-FPM  & 52.2 \\ 
    \bottomrule
    \end{tabular}
\caption{Ablation study on different setting (including number of initial sampling frames and the ratio of kept frames) of the sampling module.}
\label{tab:sampling_ablation_2}
\end{table}

\section{Visualizations of SGFS}
\label{sec:add-vis}

We visualized the sampling results of SGFS and compared the sampled frames obtained through uniform sampling, SGFS, and SGFS without the length penalty under the condition of retaining the same target number of frames. From the results, we observed that the frames sampled by SGFS contain significantly less redundancy, enabling more diverse and informative visual content with the same number of frames.

\begin{figure*}
    \centering
    \includegraphics[width=1\linewidth]{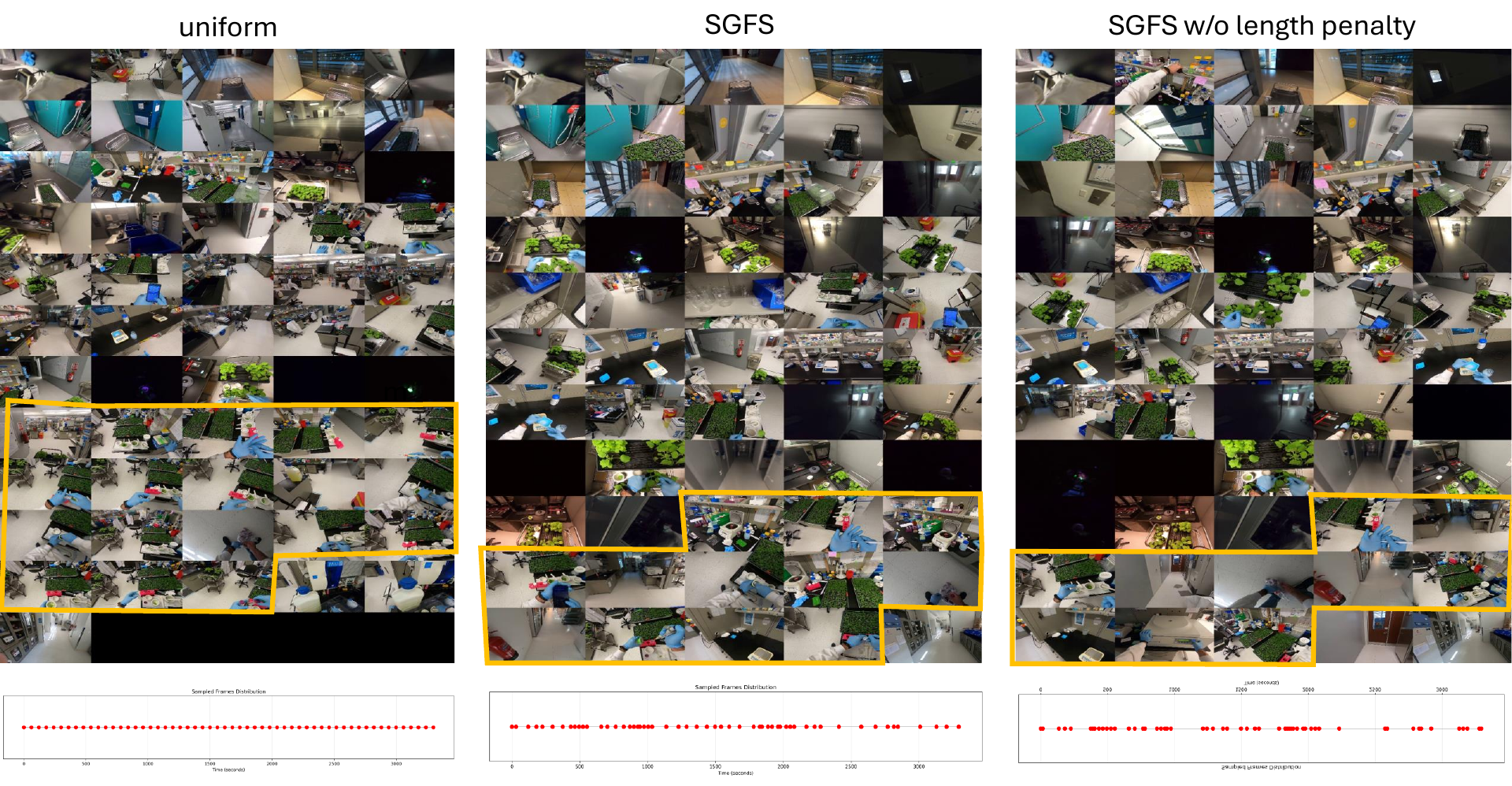}
    \caption{Sampled frames thumbnails and timeline distribution}
    \label{fig:enter-label}
\end{figure*}

\begin{figure*}
    \centering
    \includegraphics[width=1\linewidth]{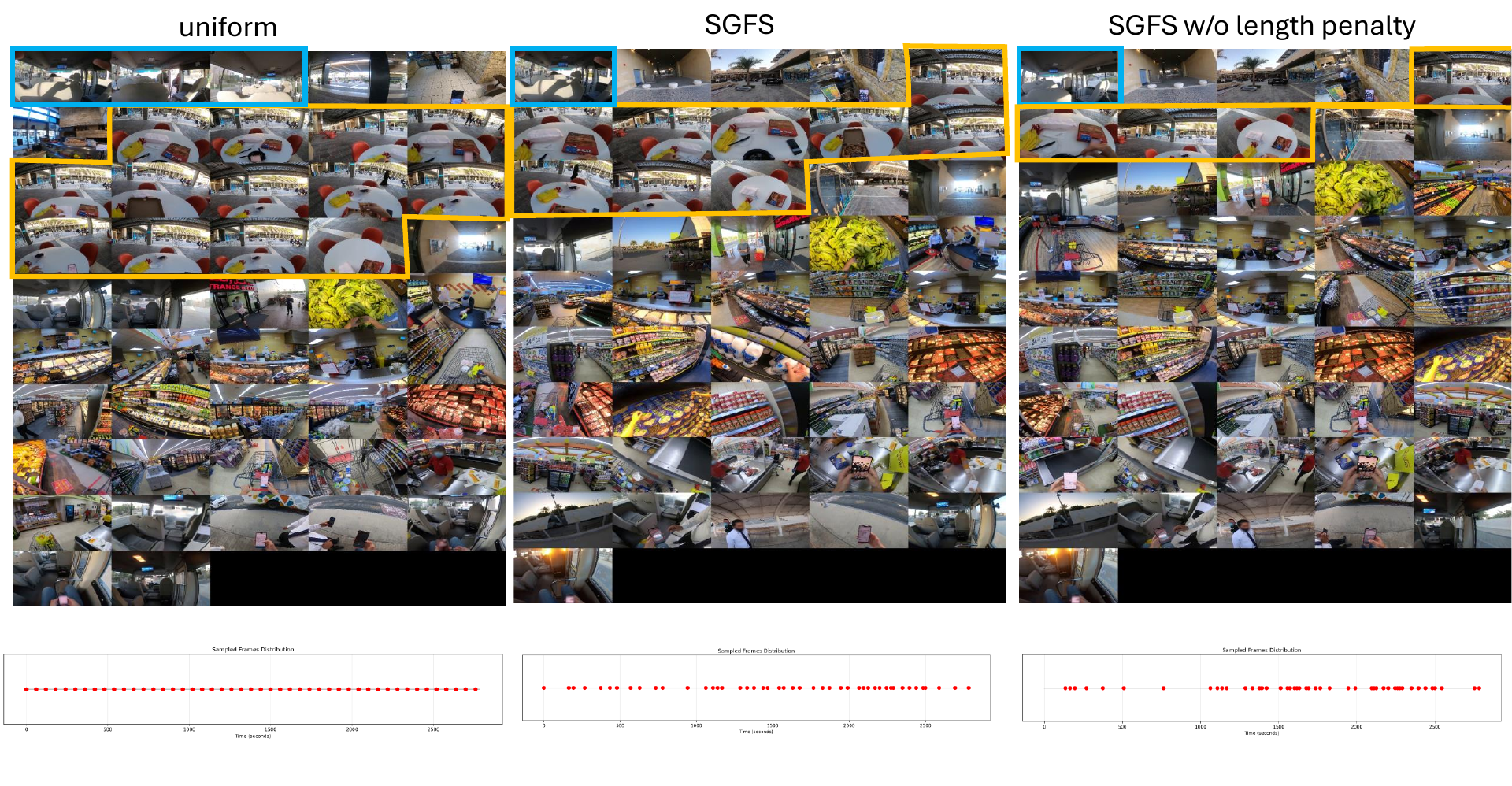}
    \caption{Sampled frames thumbnails and timeline distribution}
    \label{fig:enter-label}
\end{figure*}

\begin{figure*}
    \centering
    \includegraphics[width=1\linewidth]{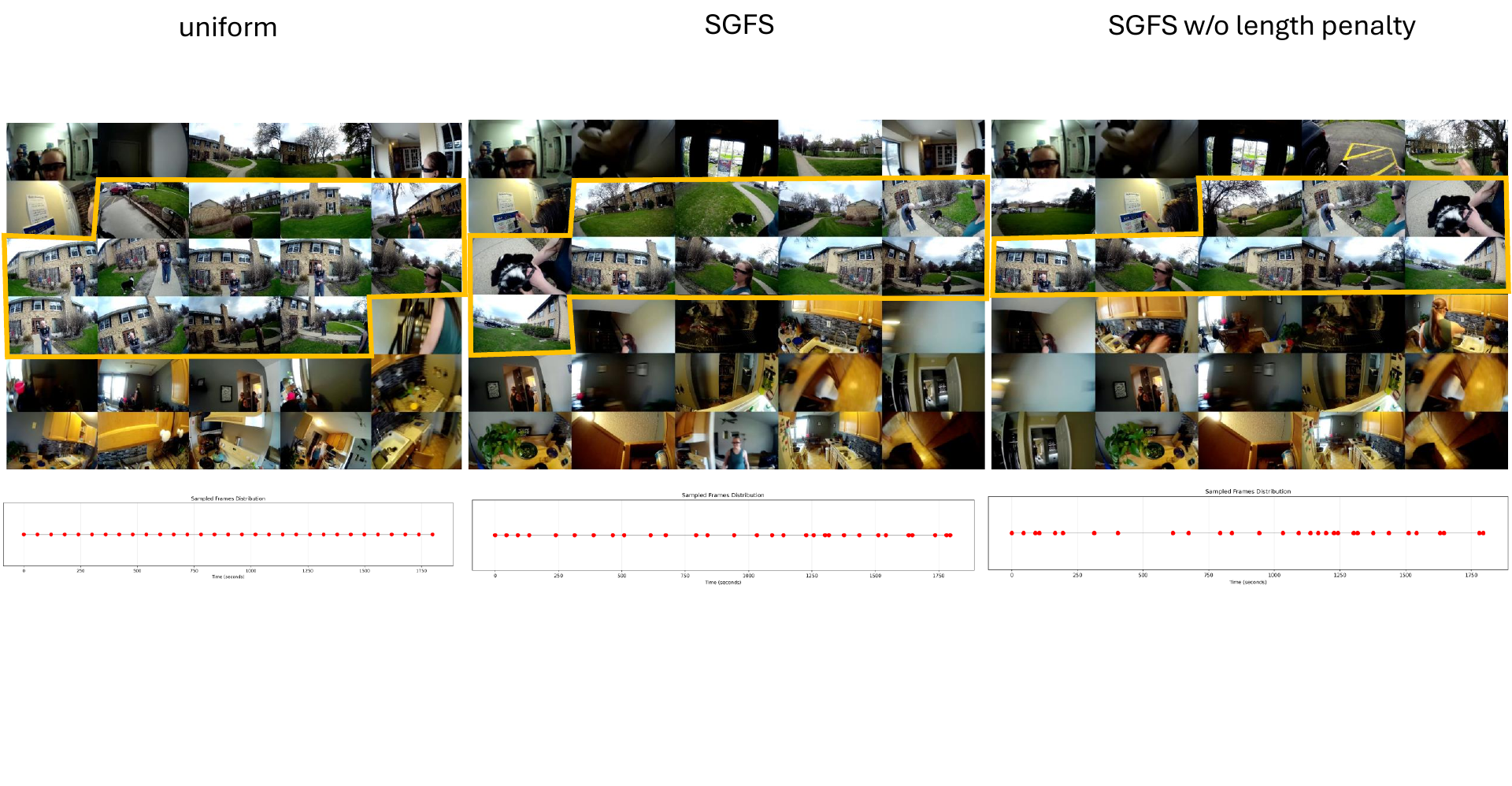}
    \caption{Sampled frames thumbnails and timeline distribution}
    \label{fig:enter-label}
\end{figure*}

\end{document}